\documentclass[10pt, conference, a4paper]{IEEEtran}
\IEEEoverridecommandlockouts
\usepackage{cite}
\usepackage{rotating}
\usepackage{amsmath,amssymb,amsfonts}
\usepackage{algorithmic}
\usepackage{graphicx}
\usepackage{textcomp}
\usepackage{xcolor}
\usepackage{flushend}
\usepackage{hyperref}
\usepackage{amsmath}
\usepackage{enumitem}
\usepackage{subcaption}
\usepackage{comment}
\usepackage[autostyle]{csquotes}
\newcommand{\linebreakand}{%
  \end{@IEEEauthorhalign}
  \hfill\mbox{}\par
  \mbox{}\hfill\begin{@IEEEauthorhalign}
}

\usepackage[pscoord]{eso-pic}

\def\BibTeX{{\rm B\kern-.05em{\sc i\kern-.025em b}\kern-.08em
    T\kern-.1667em\lower.7ex\hbox{E}\kern-.125emX}}
\begin{document}

\title{
SPARROW: Smart Precision Agriculture Robot for Ridding of Weeds
}

\makeatletter

\author{Dhanushka Balasingham$^{1}$, Sadeesha Samarathunga$^{2}$, Gayantha Godakanda Arachchige$^{3}$, Anuththara Bandara$^{4}$, \\Sasini Wellalage$^{5}$, Dinithi Pandithage$^{6}$, Mahaadikara M.D.J.T Hansika$^{7}$ and Rajitha de Silva$^{8}$
\\
\IEEEauthorblockA{$^{1-4,6-7}$ Dept. of Computer Systems Engineering, Sri Lanka Institute of Information Technology, Sri Lanka}
\IEEEauthorblockA{$^{5}$ City of Melbourne, Australia}

\IEEEauthorblockA{$^{8}$ Lincoln Agri-Robotics, University of Lincoln, United Kingdom}

\thanks{ 
        {\tt\small $^{1}$dhanushkasandeepa@gmail.com,$^{8}$rajitha@ieee.org} (Correspondence author: Rajitha de Silva)}%
}
\maketitle

\begin{abstract}
The advancements in precision agriculture are vital to support the increasing demand for global food supply. Precision spot spraying is a major step towards reducing chemical usage for pest and weed control in agriculture. A novel spot
spraying algorithm that autonomously
detects weeds and performs trajectory planning for the sprayer
nozzle has been proposed. Furthermore, this research introduces a vision-based autonomous navigation system that operates through the detected crop row, effectively synchronizing with an autonomous spraying algorithm. This proposed system is characterized by its cost effectiveness that enable the autonomous spraying of herbicides onto detected weeds.

\end{abstract}



\section{Introduction}
Agriculture, as an irreplaceable pillar of human survival and a key driver of economic growth in nations, assumes a critical role in meeting the higher demand for food and agricultural requirements. However, farmers are faced with multifaceted challenges, particularly in the realm of weed control, as they strive to ensure the long-term sustainability of the agricultural sector. Weeds that compete with crops for essential resources create a substantial threat to crop yield and quality of harvest\cite{korav2018study}. Conventional practices of manual weeding are labor-intensive, time-consuming, and financially burdensome for farmers\cite{inbook}. In response, the adoption of herbicides has gained prominence as a cost-effective solution for weed management. Nevertheless, the excessive use of herbicides gives rise to environmental concerns and potential health risks for both humans and animals. To address these challenges and enhance precision agriculture practices, the integration of Robotics and Artificial Intelligence (AI) technologies holds the promise of revolutionizing weed control methods. By developing a precise and targeted herbicide application system, it is possible to minimize the number of chemicals employed, thereby reducing environmental pollution and mitigating risks to human health\cite{Kortekamp_2011}.  The research commences by presenting statistical insights into the present-day utilization of robotics in agriculture. The statistical data presented in these papers \cite{stat1}, \cite{stat2} underscores the market dimensions of agricultural robotics and predicts its expansion in the coming years. This research paper endeavors to explore the development of a Smart Precision Agriculture Robot for Ridding Of Weeds (SPARROW) which harnesses state-of-the-art technologies to effectively address the weed control problem.

The Weed Detection and Spot Spraying Robot will serve to reduce manual labor requirements, minimize herbicide usage, and optimize crop yield, thereby ensuring the long-term viability of the agricultural sector while mitigating adverse environmental impacts.
The proposed research aims to develop a fully autonomous mobile robot that integrates computer vision, image processing, and active navigation to facilitate precision agriculture. Vision based system is a more cost-effective solution than other sensory systems such as Light Detection and Ranging (Li-DAR) and Global Positioning System (GPS). The perception system plays a pivotal role in understanding the environment, enabling accurate identification of crop rows. A newer version of the You Look Only Once (YOLO) algorithm is used to detect weeds. Developing a herbicide sprayer system by integrating the information processed by the weed detection system enables more accurate spot spraying.  Furthermore, the robot possesses the capability to detect weeds while navigating and selectively spray herbicides as necessary.
The main expected outcome of the project is to develop a fully autonomous mobile robot that is capable of detecting weeds and spraying herbicides for detected weeds. The key achievements of this study can be presented as follows:
\begin{itemize}
  \item Novel spot spraying algorithm that autonomously detects weeds and trajectory planning.
  \item A trajectory plan for spray herbicides.
  \item A vision based autonomous navigation system through the detected crop row that synchronises with an autonomous spraying algorithm.
\end{itemize}

\section{Related Work}
\label{sec:bgr}
YOLO is capable of detecting objects in real time, and it is competent to work under low processing power. In the paper \cite{10011010}, YOLOv3 is used to detect weeds in an agricultural field. They have been able to develop a model with a higher level of accuracy for weed detection. But in YOLOv3, since its object size is fixed scale, the model may ignore the different sizes of weeds. At the same time, a sprayer system would not be able to target the weeds accurately. Osorio el at \cite{2} analyzed that YOLOv3 was using much more interference time than Mask R-CNN as well.
The lightweightness of a neural network model is often influenced by the number of parameters it contains. A lightweight model is one that has a relatively small number of parameters compared to larger, more complex models. According to research conducted by Farooq el at \cite{9847812} , they proposed a lightweight system that uses YOLOv4-Tiny and CSPDarkNet53 as architecture, but it uses 27.6 million parameters. To overcome this issue, Liu Cheng et al  \cite{9727999}, have developed a weed detection system using YOLOv4 based on Mobilenetv3 architecture. It utilized 5.4 million parameters which makes it more lightweight than the previous research. Our proposed system utilizes the YOLOv8n model that uses only 3.2 million parameters which makes our approach lighter than the previous systems.
In the research, the weed detection system is utilizing YOLOv8 due to the fact that the capabilities of this version are more effective than other versions of YOLO.
In this paper \cite{8255361}, they have proposed an autonomous sprayer robot. This is a cost-effective approach, and it uses simple technologies that make it easy to use. This was mainly designed for the reduction of human-related tasks in the agriculture field and to protect farmers from the harmful effects of medicines such as herbicides and pesticides. Ultrasonic sensors are used to design the system. The main drawback of the system is selective spraying is not available. In this approach, the autonomous sprayer utilizes vision sensor information which allows it to distinguish weeds and plants, then it applies herbicides to the targeted weeds only.
Authors of \cite{10039828}, \cite{10113107} have proposed spraying systems for agricultural tasks. Both systems use LIDAR-based navigation. It is able to navigate in a robust environment with high accuracy. The main drawbacks of such a system are that LIDAR-based navigation systems are very expensive and the proposed system’s spraying mechanism is not automated. When it comes to spraying, both developed systems in \cite{10039828}, \cite{10113107} need human interaction through a mobile application in order to function the sprayer system. In this research, vision-based navigation was implemented because that significantly reduces the cost of the robot compared to LIDAR-based and GPS-based systems. The methodology outlined in this study \cite{10089591} employed vision-based navigation for herbicide spraying. The proposed design required a high processing unit to accomplish the vision-based navigation effectively. Additionally, in SPARROW, the sprayer system does not require any human interaction, and it is capable of autonomously targeting weeds and spraying herbicides with low processing power.

\section{Methodology}
\label{sec:dt}
The proposed system consists of software components namely Robotic Operating System (ROS), Python, YOLO algorithms, and image processing techniques. In terms of hardware, a differential drive robot was designed with a precise herbicide sprayer and cameras placed separately to spot weeds and find crop rows. The system comprises four key subsystems that work in synergy to achieve autonomous weed detection and spot spraying of herbicides as follows.
\begin{figure}
    \centering
    \includegraphics[width=1\linewidth]{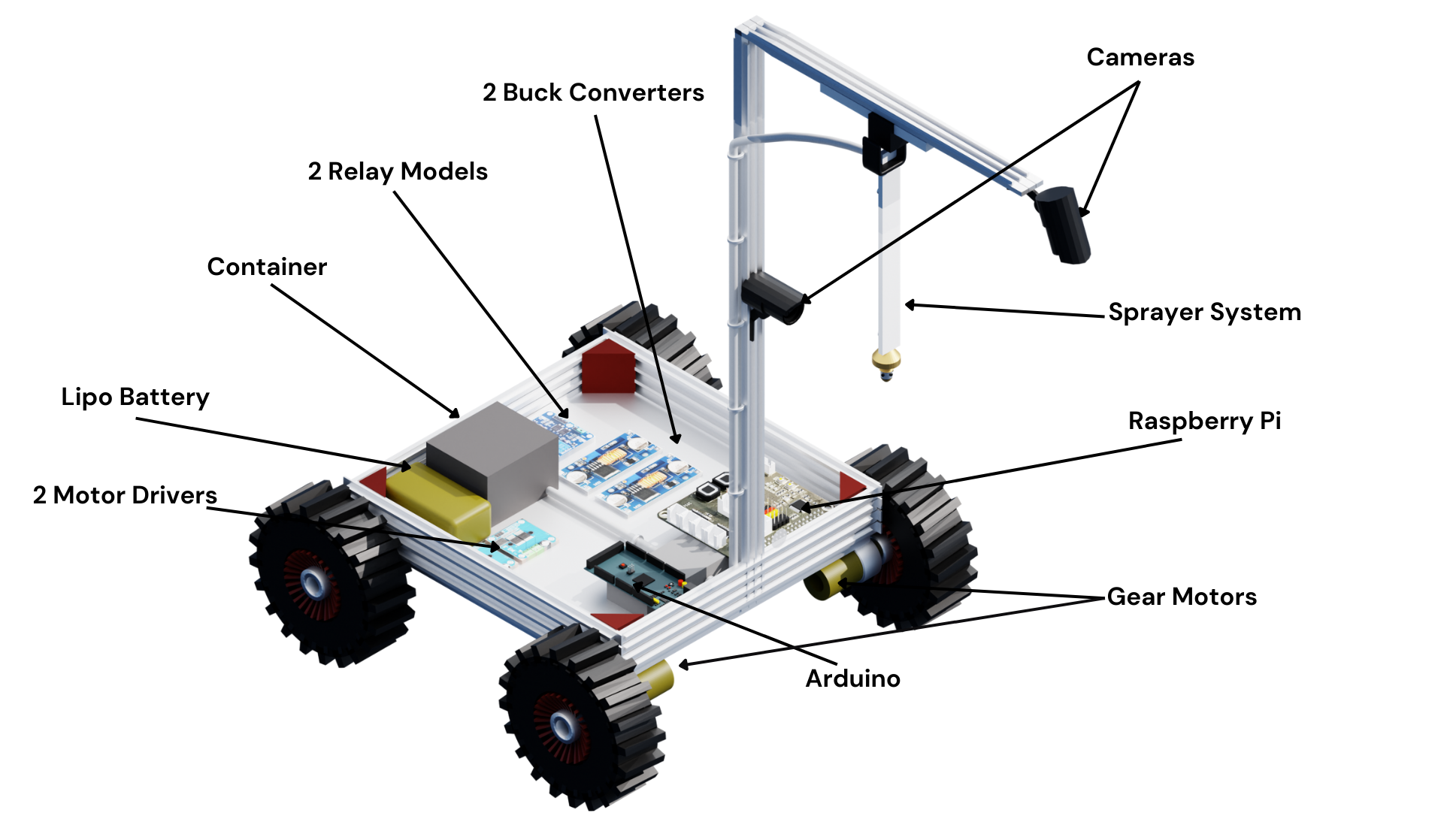}
    \caption{Design of the SPARROW.}
    \label{fig:rover}
\end{figure}

\begin{figure*}[ht]
    \centering
    \includegraphics[width=1\linewidth]{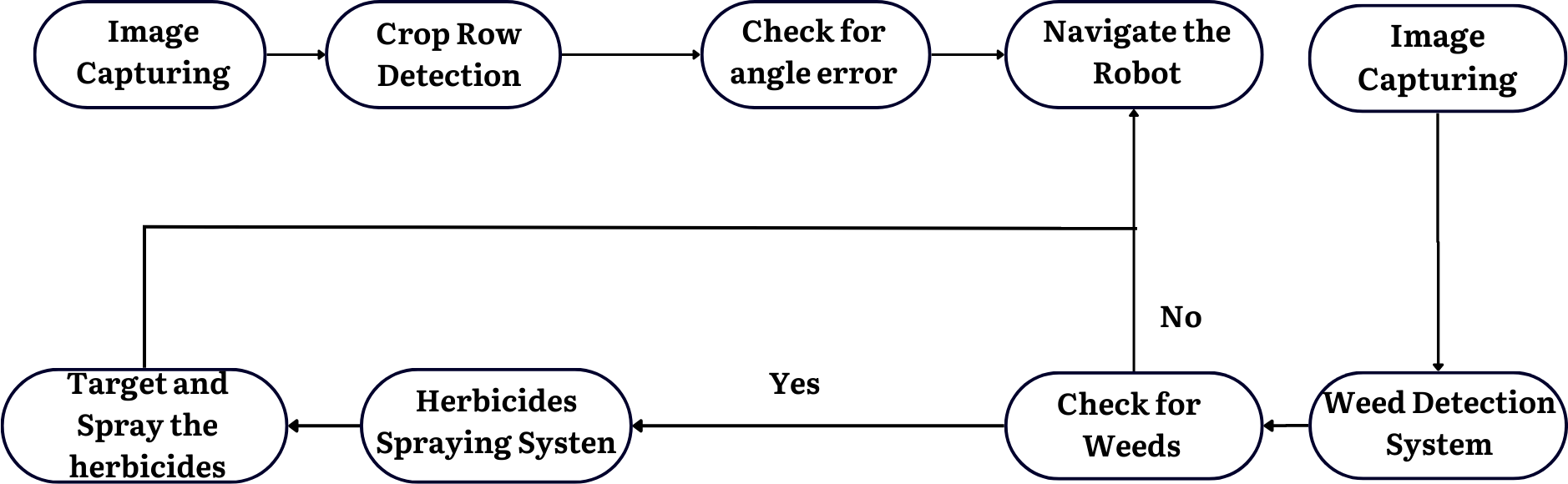}
    \caption{System diagram of the SPARROW.}
    \label{fig:system}
\end{figure*}

 Figure \ref{fig:rover} shows the design of the SPARROW robot. The SPARROW implementation cost totals around \$350. It
is powered by a 12V lithium-polymer battery, providing approximately 1.5 hours of operation. The entire system is integrated by using Robotic Operating System (ROS) middleware and the infrastructure is combined using ROS nodes and topics. These subsystems include a vision-based perception system, a vision-based autonomous navigation system, a weed detection system, and a herbicide sprayer system. Figure \ref{fig:system} illustrates the overall architecture of the proposed system.

\subsection{Vision-Based Perception System}
Understanding the environment and detecting the crop row is the initial step in the proposed active navigation perception-action loop. A vision-based approach was used to perceive the environment. A wide-angle webcam is used to capture the perceived data. The webcam resolution is 720p at 30fps. Also, it has a 100-degree wide-angle coverage. The front-mounted camera in the robot is able to capture multiple crop rows. The dataset was generated by recording videos from the robot's perspective, which were subsequently divided into individual frames. After collecting the dataset, image processing techniques that are used in this research \cite{9636267} such as Gaussian blur, RGB color filtering, erosion and dilation, canny edge detection, and region of interest cropping are applied to segment the crop rows. Firstly, Images were resized and later Gaussian blur filter was applied to reduce the noise and weeds in the path. In the next step, the image was split into RGB color channels. Normalized Difference Index (NDI) \cite{MEYER2008282} equations are also used in this process. Afterward, performing Morphological filters such as erosion and dilation with necessary iterations could be used to get a smooth contour. In such a way, the system would be able to identify the crop rows accurately. Afterward by utilizing the triangle scanning algorithm \cite{DESILVA2024108581}, the system is able to point out the crop rows in the image. After determining the middle of the crop row, it needs to calculate the error of the angle with respect to the current position.

\subsection{Vision-based autonomous navigation system}
The robot's autonomous motion-controlling part can be achieved using the vision-based autonomous navigation system. Navigation system act as a centralized system for other systems. Upon weed detection, the weed detection system generated an interrupt signal. Consequently, the navigation was temporarily suspended. Following herbicide application, the sprayer system generated a resume signal that allows the navigation process to resume. Also, the perception system and navigation system are interconnected. Through their collaborative efforts, the proposed system attains vision-based navigation across the crop row. The approach relies on local imagery features and doesn't require the use of an explicit field map. A proportional controller was implemented to steer the robot to follow the central crop row\cite{DESILVA2024108581}. The linear velocity is consistently maintained at a constant level and the angular velocity ($\dot{\omega}$) is given in the equation \ref{eq:one} while the $\alpha$ is the proportional gain and $\Delta\theta$ is the angle error. 
\begin{equation} \label{eq:one}
  \dot{\omega} = \alpha \left( \Delta\theta\right)
\end{equation}

\subsection{Weed detection system}
Prior to spraying herbicides to the weeds, the SPARROW needs to identify the weeds. The weed detection system is developed using YOLO version 8. YOLOv8 uses a c2f module which reduces the computational complexity and helps to maintain a lightweight design \cite{app13158890}. So, YOLOv8 can detect weeds more quickly and accurately than previous versions of the algorithm \cite{unknown2}\cite{unknown3}. This will ensure that the robot works real time and detects only weeds while reducing the risk of accidentally spraying herbicides on other crops. It uses a new feature pyramid network (FPN) called FPN+PAN (Pyramid Attention Network), which is more effective at detecting objects of different sizes and scales \cite{unknown3}. This is important for weed detection because weeds can vary in size, and the model needs to be able to detect them regardless of their size. So, this allows the model to detect weeds even if they are small or obscured by other objects. 

A separate camera is used for weed detection. This camera is set up on the front pole and is pointed towards the ground, covering the 100-degree angle that identifies the stems of both crop rows. It helps to detect weeds in the path as well as weeds located near the crops. This form of detection holds considerable significance, as the paper \cite{XI202220} defines that weeds could be positioned either around the middle of the crop row or in proximity to the crop row. Furthermore, the weed detection model has been executed on a Raspberry Pi 4 model. The images captured were of the weeds that are in the experimental field. The dataset of weed consists of 1000 images and the respective labels. CVAT which is an open source web application for data annotation was used to label the data. During the data collection process, special emphasis was taken to ensure that the collected data enables the model to distinctly differentiate between weeds and crops. Since the robot required weed detection in real-time, the system is developed using the YOLOv8 nano model which is the fastest and requires only a small amount of processing power than other models. The object detection method is used to detect the weeds. When weeds are detected, firstly it is going to send an interrupt signal to the navigation system which will stop the navigation until the spray is done and at the same time, the model will draw a bounding box around the weed with its coordinates.
\subsection{Herbicides Sprayer System}
\begin{figure}[t]
    \centering
    \includegraphics[width=1\linewidth]{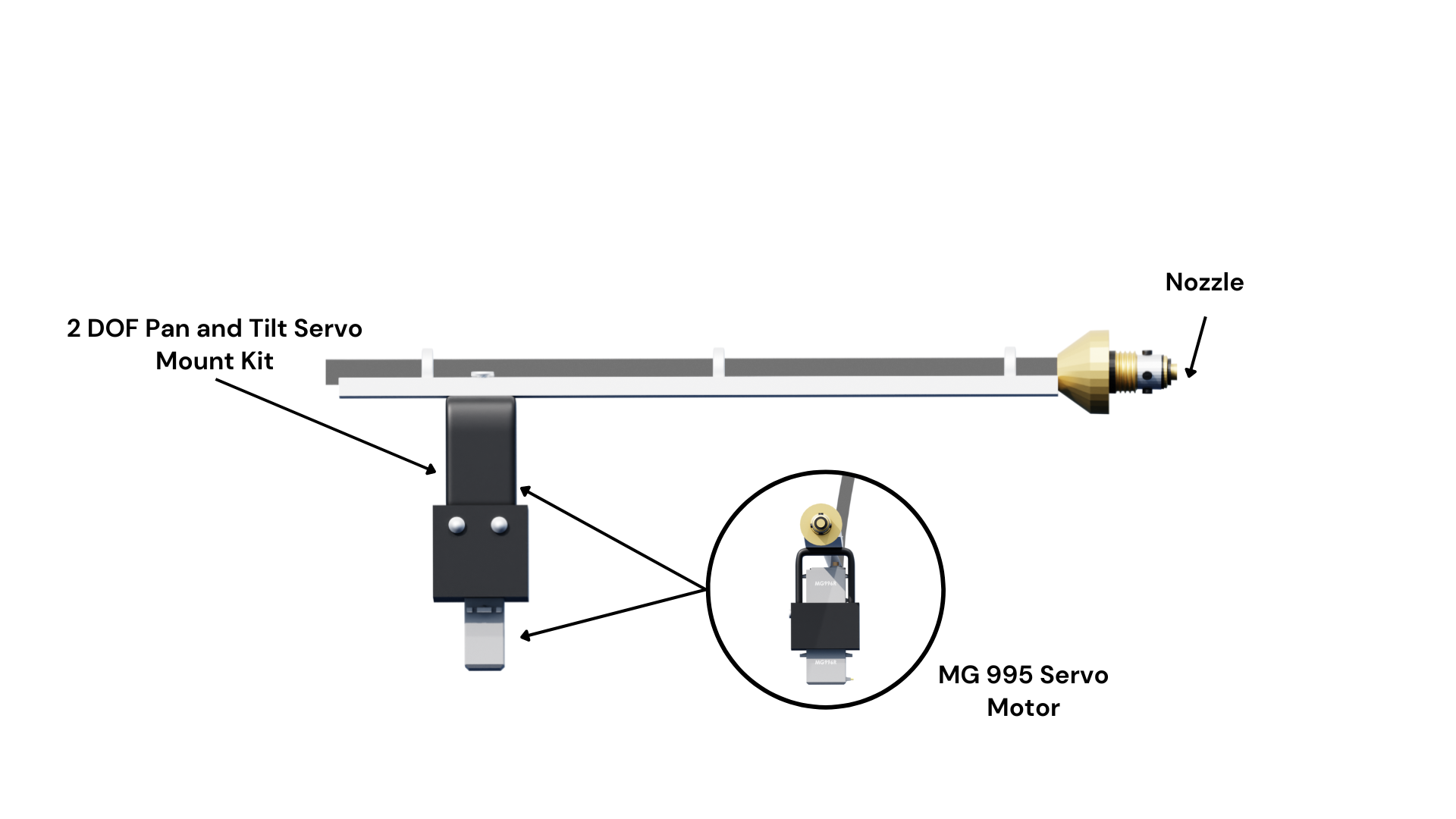}
    \caption{Design of the herbicides sprayer.}
    \label{fig:sprayerS}
\end{figure}

The herbicide spraying system is dedicated to spraying herbicides to the detected weeds accurately which is the most critical task in the agricultural environment. Figure \ref{fig:sprayerS} shows the design of the herbicide sprayer system of the SPARROW.

The sprayer system is implemented using the integration of two trajectory planning algorithms. These algorithms collectively determine the most efficient path for the sprayer to traverse once weed coordinates are detected. In both methods, the initial reference point is set as the central point within the camera's coverage area which is described in Section \ref{sec:herspp}. The first method is the nearest neighbour algorithm which is used to determine the sprayer trajectory path by selecting the closest weed. The second method is the Christofides algorithm \cite{turret1} which is used to calculate an optimal path for the sprayer system to follow. This path ensures that the turret visits each weed location while minimizing the overall distance traveled. The algorithm helps optimize the sequence in which the sprayer should move to eliminate the weeds efficiently. 

\section{Results and Discussion}
\label{sec:rslt}

\subsection{Evaluation of Weed Detection System}
\label{sec:wd}

\begin{figure}
    \centering
    \subfloat[]{\includegraphics[width=0.24\textwidth, height=4cm]{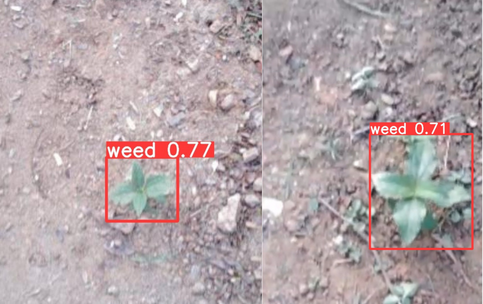}}\hfill
    \subfloat[]{\includegraphics[width=0.24\textwidth, height=4cm]{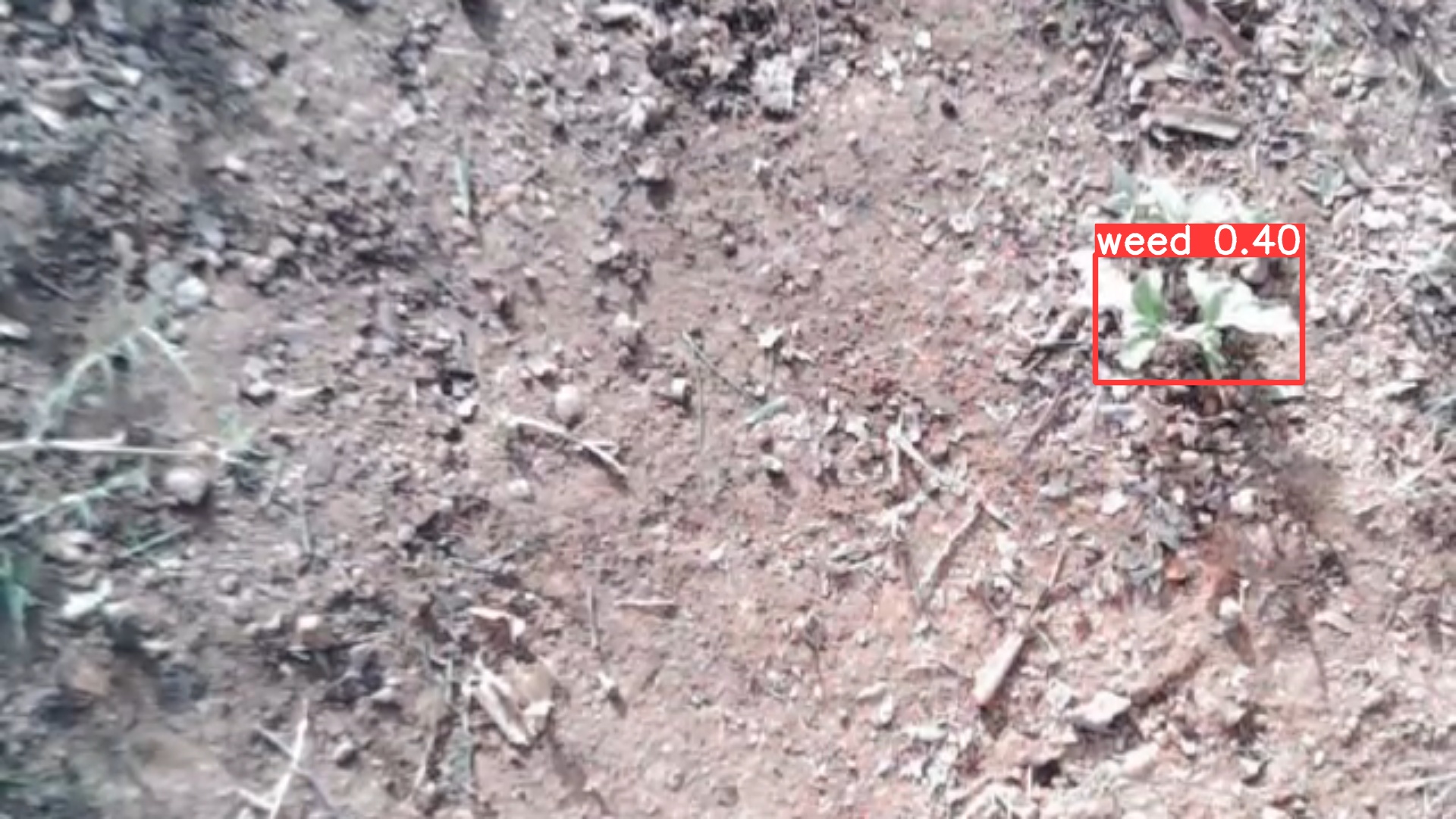}}\hfill
    \subfloat[]{\includegraphics[width=0.24\textwidth, height=4cm]{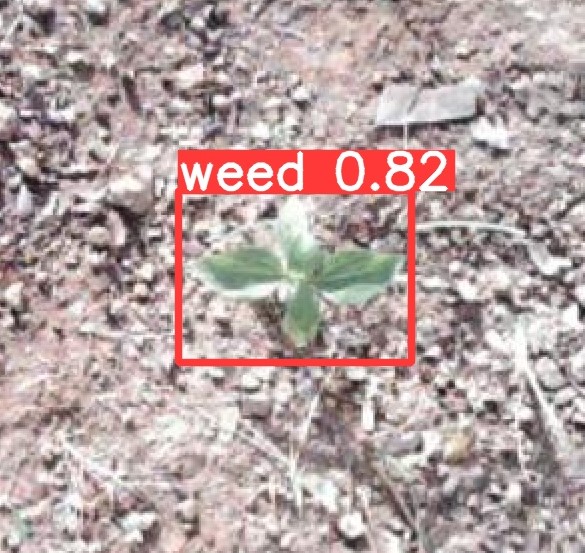}}\hfill
    \subfloat[]{\includegraphics[width=0.24\textwidth, height=4cm]{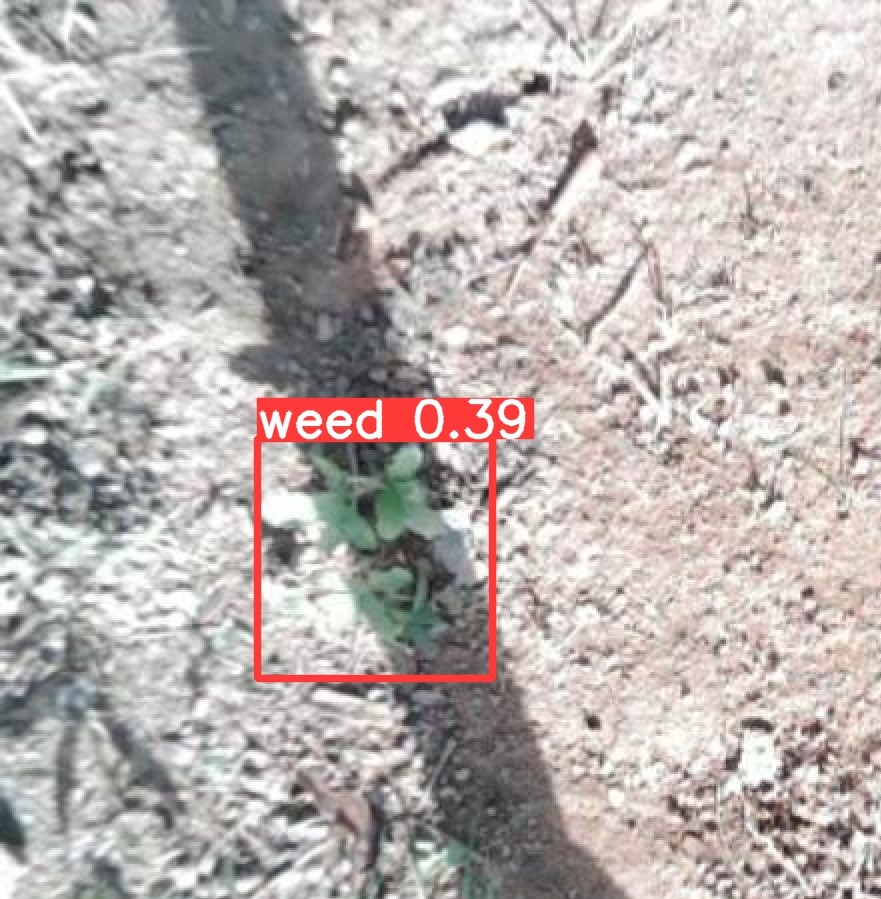}}\hfill
    \caption{Output of weed detection system. a: Typical condition, b,c:  Sunny condition, d: Shadowed condition.}
    \label{fig:weedDet}
\end{figure}

\begin{figure}
    \centering
    \includegraphics[width=1\linewidth, height=7.5cm]{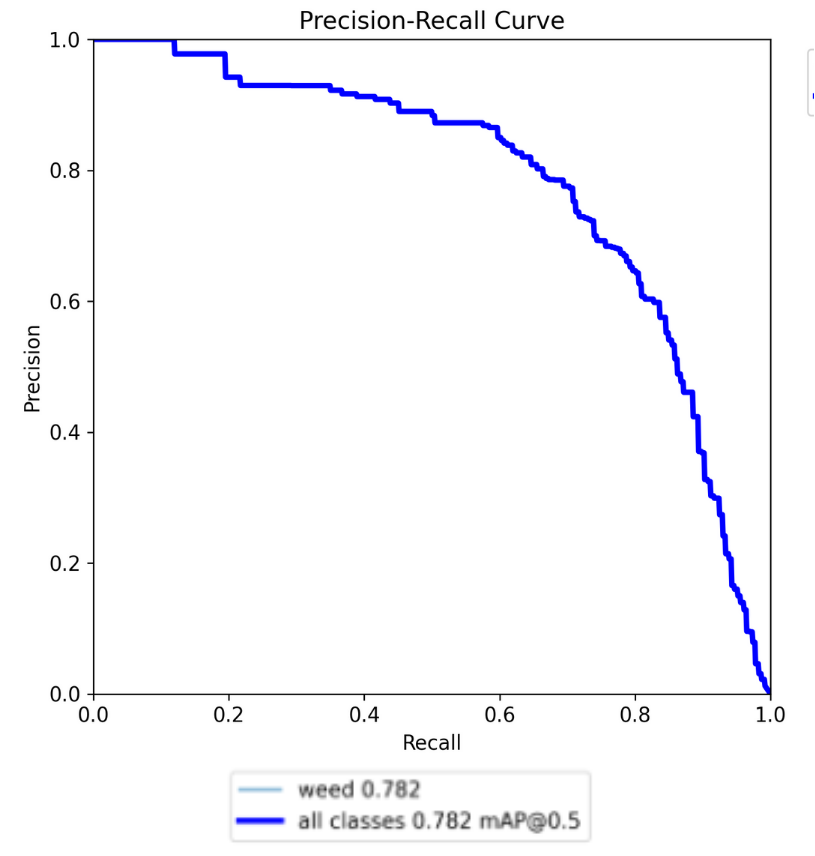}
    \caption{Precision-Recall curve.}
    \label{fig:pr}
\end{figure}

\begin{figure}
    \centering
    \includegraphics[width=1\linewidth]{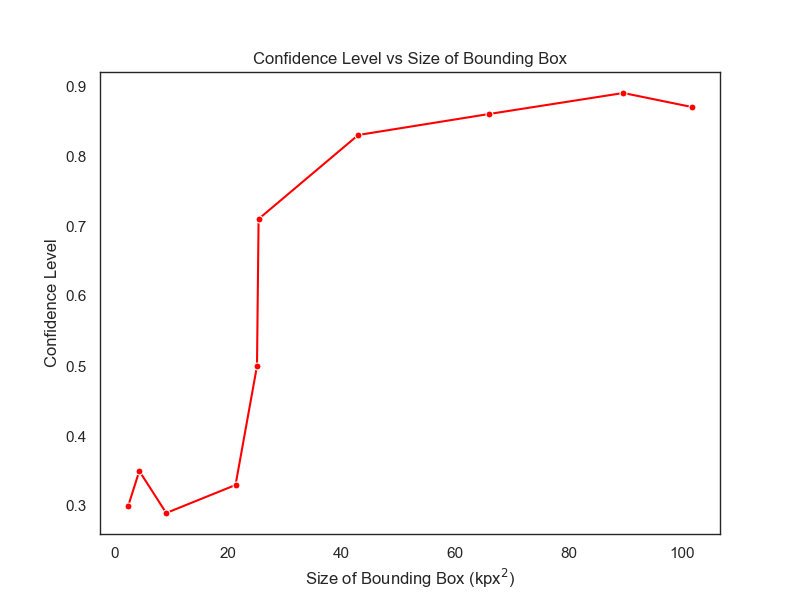}
    \caption{Relationship between size of bounding box and confidence level.}
    \label{fig:wc}
\end{figure}

The precision-recall curve is used as an evaluation metric to gauge the effectiveness of the weed detection algorithm. In the precision-recall curve depicted in Figure \ref{fig:pr}, the y-axis represents precision, indicating the proportion of correctly predicted positive detection among all predicted positives. Meanwhile, the x-axis represents recall, measuring the algorithm's proficiency in capturing actual positive instances. The calculated Area Under the Precision-Recall Curve (AUC-PR) further validates the algorithm's superior performance. The graph shown in Figure \ref{fig:wc} demonstrates the correlation between confidence levels and the predictive performance of the weed detection system across a range of bounding box sizes. As indicated by the graph, the system performs a high level of confidence when detecting larger-sized weeds. Furthermore, the system is capable of detecting weeds under multiple conditions as shown in Figure \ref{fig:weedDet} which ensures the system performs effectively in diverse and robust environmental settings.

\subsection{Evaluation of trajectory planning}
\label{sec:hersp1}

The performance of trajectory planning by utilizing the nearest neighbour algorithm is evaluated by the $\Phi_{N}$ score described in equation \ref{eq:nn}. The $\Phi_{N}$ score is defined using in terms of $\lambda_N$ and $\lambda_O$. The $\lambda_O$ represents the distance of the most optimal path, while $\lambda_N$ signifies the calculated distance determined by the nearest neighbor algorithm. The Optimal-to-Nearest Ratio ($\Phi_{N}$) is then derived from these two values. The nearest neighbour algorithm produced an average $\Phi_{N}$ score of 93.99\% across the entire dataset.

\begin{equation} \label{eq:nn}
\Phi_N = \frac{\lambda_O}{\lambda_N}\
\end{equation}

The evaluation of trajectory planning using the Christofides algorithm is assessed through the $\Phi_C$ score, as outlined in equation \ref{eq:chr}. The $\Phi_C$ score is defined in relation to $\lambda_C$ and $\lambda_O$. Here, $\lambda_C$ represents the calculated distance determined by the Christofides algorithm. Subsequently, the Optimal-to-Christofides Ratio ($\Phi_C$) is computed based on these two values. Finally, Christofides algorithm produced an average of $\Phi_C$ score of 93.22\% in the entire dataset. 

\begin{equation} \label{eq:chr}
\Phi_C = \frac{\lambda_O}{\lambda_C}\
\end{equation}

\begin{figure}[ht]
    \centering
    \includegraphics[width=1\linewidth]{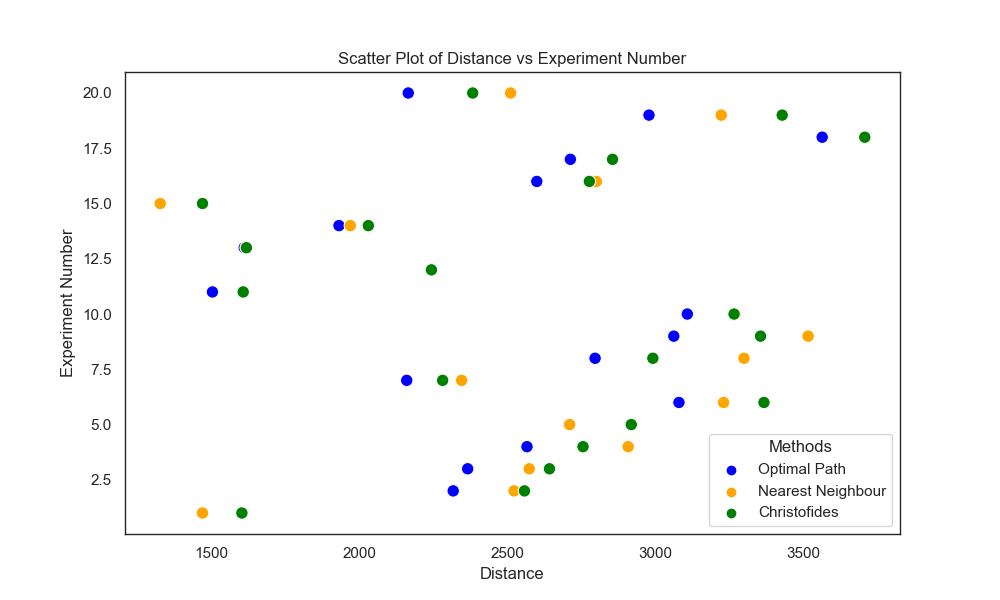}
    \caption{Performance of the algorithms over distance.}
    \label{fig:expr}
\end{figure}

\begin{figure}
    \centering
    \includegraphics[width=1\linewidth]{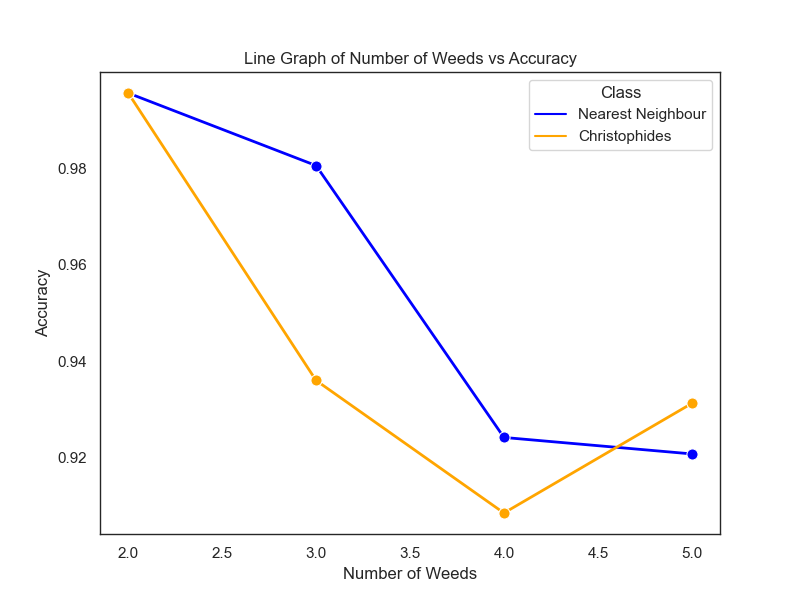}
    \caption{Performance of the algorithms over number of weeds.}
    \label{fig:resT}
\end{figure}

Figure \ref{fig:expr} illustrates the performance outcomes of the trajectory algorithms. Figure \ref{fig:resT} visualizes the algorithms' accuracy in relation to the number of captured weeds. These results depict that when the weeds are increased, the Christophides algorithm performs better than the nearest neighbour algorithm. Hence, the nearest neighbour algorithm proves to be more suitable for conditions where the number of weeds is relatively lower.

\subsection{Placement of herbicides sprayer}
\label{sec:herspp}

An experiment was set up to determine the optimal positioning of herbicides sprayer. The objective of the experiment was to determine the maximum distance at which the sprayer system could accurately deliver liquid to the intended target. 

\begin{figure}
    \centering
    \includegraphics[width=1\linewidth]{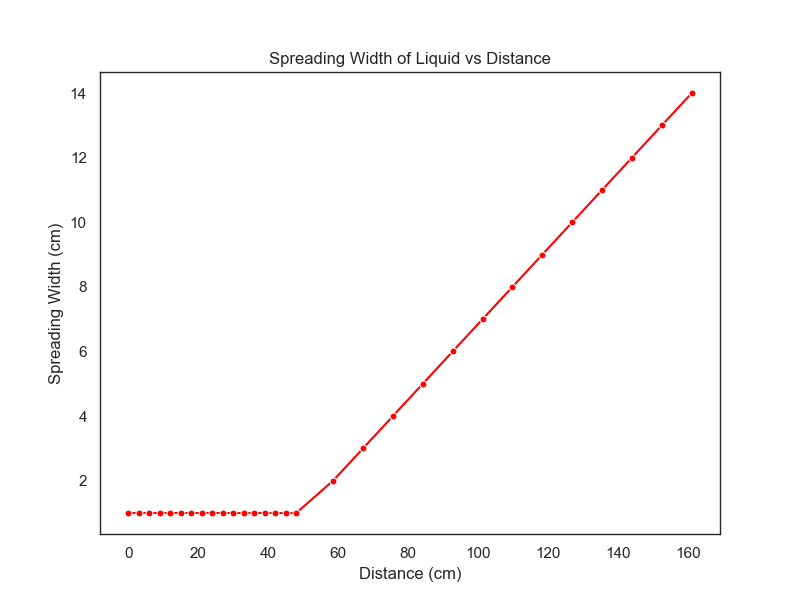}
    \caption{Relationship between spreading width of liquid and distance.}
    \label{fig:resherbS1}
\end{figure}

\begin{figure}
    \centering
    \includegraphics[width=1\linewidth]{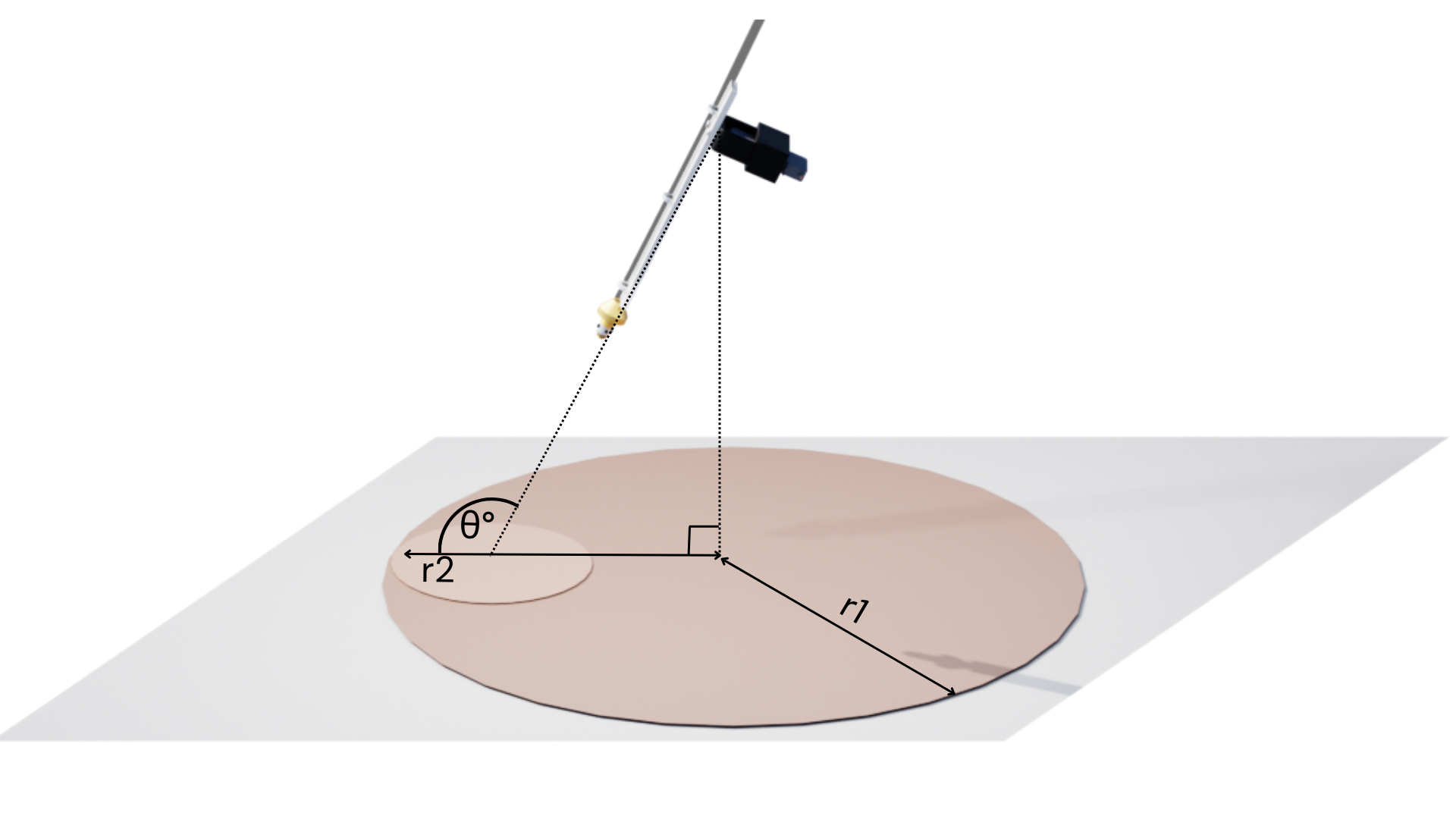}
    \caption{Placement of sprayer.}
    \label{fig:resherbS2}
\end{figure}

The sprayer was mounted at a height of 70 cm above ground. The sprayer is facing the ground plane at $\theta=90$\textdegree in its normal position and the placement of the sprayer is shown in Figure \ref{fig:resherbS2}. The ground distance at which the sprayer was aimed away from its normal position was characterized by spray radius $r_1$. The radius of the area which the sprayed liquid covers is identified by target radius $r_2$. The sprayer was then gradually moved away from the normal position while increasing $\theta$ above $90$\textdegree. The sprayer could achieve a minimum $r_2$ of 1 cm with an $r_1$ up to 50 cm from the normal position. The $r_2$ increased linearly up to a maximum of 14 cm at a maximum $r_1$ of 161 cm. The camera employed for weed detection can cover an area of 96cm x 51cm within the field. Thus, this experiment deduced that for effectively applying herbicides to weeds situated along the boundaries, the sprayer system must be positioned at the midpoint of the horizontal pole. Additionally, the nozzle should be directed towards the center of the area captured by the camera.

\subsection{Evaluation of Perception System}

\begin{figure} 
    \centering
    \subfloat[]{\includegraphics[width=0.16\textwidth, height=4cm]{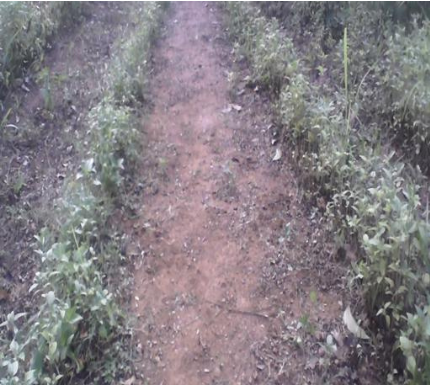}}\hfill
    \subfloat[]{\includegraphics[width=0.16\textwidth, height=4cm]{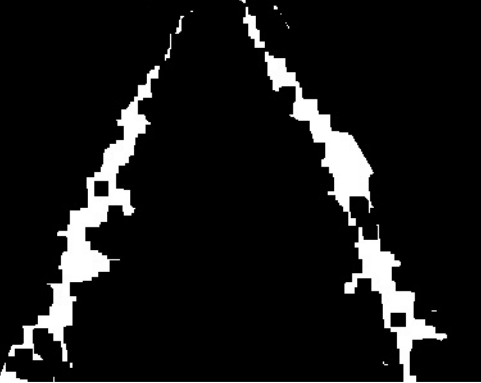}}\hfill
    \subfloat[]{\includegraphics[width=0.16\textwidth, height=4cm]{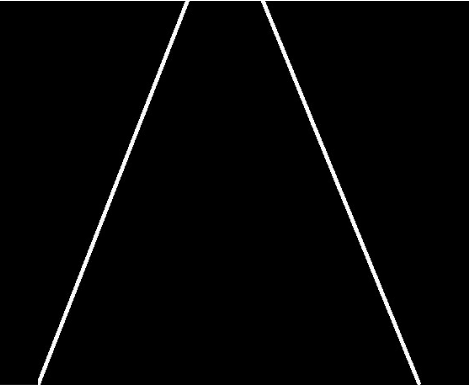}}
    \caption{Stages of crop row detection. a: Input RGB image, b: ExG Image, c: Ground truth image.}
    \label{fig:foobar}
\end{figure}

The evaluation of crop row detection is carried out using the Intersection Over Union (IOU) score. Figure \ref{fig:foobar} shows the process of identifying the crop rows. The average IOU score for the  dataset in this method was 39.39\%. The IOU score is calculated by comparing the image-processed data (b) with the corresponding ground truth label (c) in Figure \ref{fig:foobar}. Since the IOU value is influenced by the shape of the segmented data, it's important to note that crops, being thin and narrow objects, may result in an IOU value that indicates a certain level of error. For this reason, even though the system performed a lower score, this error could be regarded as a minor offset.

\section{Conclusion}

The proposed system is capable of autonomously navigating through crop rows, simultaneously detecting weeds, and applying herbicides to the identified weeds. The weed detection system utilized the YOLOv8 algorithm and it scored higher AUC-PR value. Also, the weed detection system works in a diverse and robust environment efficiently. Trajectory planning algorithms are based on the concepts of nearest neighbour and Christofides algorithms. These algorithms achieved  average scores of 93.99\% and 93.22\% for $\Phi_{N}$ and $\Phi_{C}$ respectively. Consequently, an algorithm has been developed that integrates elements from both approaches. The system employed the nearest neighbour algorithm for a lesser number of weeds present. Christofides algorithm is utilized when they have a larger number of weeds and a more extensive distribution of weeds across the field. The perception system is capable of detecting the crop row with a score of 39.39\% IOU value which could be considered a promising score to detect crop row in such an agricultural environment. The vision-based approach is indeed more cost-effective compared to Lidar and GPS-based systems.  The proposed vision-based approach is indeed more cost-effective compared to Lidar and GPS-based systems.

\bibliographystyle{./IEEEtran}
\bibliography{root}
\end{document}